\documentclass[letterpaper, 10 pt, conference]{ieeeconf}  %

\usepackage{cleveref}
\usepackage{float}
\usepackage{graphics} %
\usepackage{graphicx}
\usepackage{subcaption}
\usepackage{amssymb}
\usepackage{url}
\crefname{figure}{Fig.}{Figs.}
\Crefname{figure}{Fig.}{Figs.}

\IEEEoverridecommandlockouts                              %

\title{\LARGE \bf
Robotic Optimization of Powdered Beverages Leveraging Computer Vision and Bayesian Optimization
}%

\author{Emilia Szymańska$^{1,3}$, Josie Hughes$^{2,3}$%
\thanks{$^{1}$ETH Z\"{u}rich, Z\"{u}rich, Switzerland; \textit{emilia.szymanska.eka@gmail.com}}%
\thanks{$^{2}$CREATE Lab, EPFL, Lausanne, Switzerland}%
\thanks{$^{3}$This work was conducted at EPFL CREATE Lab in partnership with Nestle Research, based in Lausanne, Switzerland.}
}

\begin{document}

\maketitle
\thispagestyle{empty}
\pagestyle{empty}

\begin{abstract}

The growing demand for innovative research in the food industry is driving the adoption of robots in large-scale experimentation, as it offers increased precision, replicability, and efficiency in product manufacturing and evaluation. To this end, we introduce a robotic system designed to optimize food product quality, focusing on powdered cappuccino preparation as a case study. By leveraging optimization algorithms and computer vision, the robot explores the parameter space to identify the ideal conditions for producing a cappuccino with the best foam quality. The system also incorporates computer vision-driven feedback in a closed-loop control to further improve the beverage. Our findings demonstrate the effectiveness of robotic automation in achieving high repeatability and extensive parameter exploration, paving the way for more advanced and reliable food product development.

\end{abstract}

\vspace{-0.08cm}
\section{INTRODUCTION}

    Food science is starting to play a significant role in the worldwide search for life improvements. Considering the growth of the human population and the dangers of human-induced pollution, the need to move towards more sustainable and efficient food sources is of utmost priority. To address this challenge, it is essential that these products are obtained in an optimized manner. Moreover, they must satisfy end-users’ sensory perceptions to ensure widespread acceptance and adoption~\cite{kobayashi2015impact,kang2022effects}. Research carried out in this sector requires high repeatability and accuracy in experiments whose main objective is to understand complex physical and chemical reactions occurring in food or drink preparation~\cite{juriaanse2006challenges,tuorila2009sensory, buttriss2013food}. %
    
    Investigation of these aspects is typically achieved through manual lab experiments which can be slow and costly because of the extensive exploration of experimental parameters and conditions~\cite{arteaga1994systematic}. Due to the challenges resulting from high stochasticity, automation is necessary to be applied to as many aspects as possible to reduce the influence of external factors on the preparation process. Additionally, the experimental food and drink preparation should mimic human behavior, which can be challenging as it requires both sensory and physical interactions with the product~\cite{tuorila2009sensory}. One plausible solution for automating such scientific experiments is offered by a combination of robotics with computer vision and optimization~\cite{duong2020review}, allowing for precise repetition, human behavior simulation, and intelligent data capture and analysis~\cite{khan2018towards,iqbal2017prospects}.
    
    Research exploring robots' utility in a kitchen \cite{beetz_robotic_2011, danno_cooking_2021, bollini_bakebot_nodate, beet2011robotic, bollini2011bakebot, satici2016coordinate,ilic2023complexity} still faces a multitude of challenges, especially in the area of sensory perception. In the systems employing robots to analyze and optimize food, a variety of evaluation solutions have been implemented -- user feedback~\cite{junge_improving_2020}, salinity sensors~\cite{sochacki2021closed} or tactile assessment~\cite{sochacki2021compliant, scimeca2019nondestructive}. Whilst computer vision has been investigated for use in the food processing and food science industry~\cite{ma2016applications,kulcu2018computer,deotale2020foaming}, there has been limited exploration of the use of computer vision as a means of providing rapid feedback into the food optimization process. This could assist in enabling large scale, fully automated optimization of food products and their making processes with non-invasive and cheap sensing mechanisms. 
    
\begin{figure*}[h]
        \centering
        \includegraphics[width=0.9\textwidth]{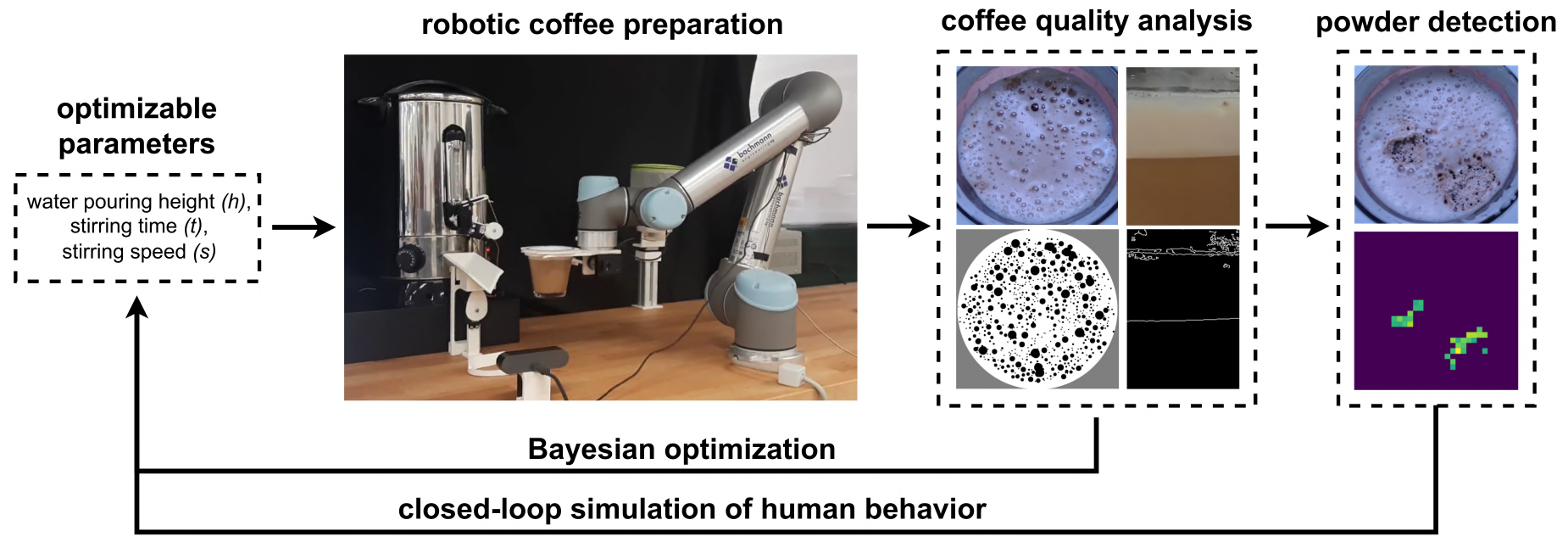}
        \caption{\textbf{Robot Food Scientist}. The robot setup with integrated computer vision is used to optimize the parameters of the beverage preparation, and to simulate human behavior in response to detection of undissolved powder clumps.}
        \label{fig:teaser}
    \end{figure*}

    To address this goal, we propose \textit{Robot Food Scientist} -- a robotic system which can automatically prepare beverages with various input parameters, evaluate their quality and optimize their creation, as presented in \cref{fig:teaser}. The selected case study is powdered cappuccino, with the foam being regarded as the primary quality indicator. We leverage computer vision analysis of the foam to simulate human responses to the visual characteristics of the beverage, with a particular emphasis on detecting and removing undissolved powder clumps in the closed-loop control system. Additionally, we use model-free optimization methods to find the optimal process for reconstitution. For foam-based beverages optimization we propose applying Bayesian Optimization (BO)~\cite{nguyen2019bayesian}, suited for sequential analysis and global optimization of black-box function without assumptions on any functional forms~\cite{shahriari2015taking}. The use of automation allowed for experiments with a high repeatability and also for much larger exploration of the different parameter combinations. 

    In summary, the paper makes the following contributions:

    \begin{itemize}
        \item We introduce the application of robotics and computer vision to conduct large-scale experiments in beverage preparation and quality analysis in an automated manner. Our case study on powdered cappuccino demonstrates the effectiveness of this approach, with over a hundred coffees prepared and systematically evaluated. 
        \item We define a set of adjustable system parameters and meaningful quality metrics for foamed beverages, and we analyze of the relationships between metrics.
        \item We develop a closed-loop control system allowing for the automated detection and removal of undissolved powder clumps, successfully mimicking human behavior in addressing product imperfections.
        \item We prove that Bayesian Optimization is an effective method for identifying optimal preparation parameters that lead to the highest quality in consumable products.
    \end{itemize}

\section{RELATED WORKS}

    \subsection{Culinary Robotics}

    Robots, as a means of addressing repetitive tasks, present a promising solution for food preparation automation. In the culinary domain, there has been a noticeable increase in research on applying robotic solutions for non-trivial tasks such as correctly tossing a pizza dough~\cite{Satici2022}, preparing a stir-fry dish~\cite{liu2022stirfry}, or making an omelet from scratch~\cite{junge_improving_2020}. Various approaches have been proposed for the implementation of these complex robotic systems. Techniques include demonstration learning~\cite{eng4040143}, integrating Internet of Things into the system~\cite{Zhao2015}, or employing Large Language Models to fulfill recipe instructions and monitor the state of food~\cite{sakib2024from, Kawaharazuka2024continuous}.

    In this study, we define a fixed set of robot sequences and enhance the execution of these sequences using computer vision as a feedback mechanism from the environment. The selected case study involves the preparation of powdered cappuccino, where hot water is added while mixing to achieve reconstitution. This process results in the dissolution of the powder into the liquid and the creation of foam through aeration. To the best of our knowledge, this is the first application of robotics combined with computer vision for the preparation and analysis of a reconstituted beverage.

    A notable challenge in the reconstitution process is the potential formation of undissolved powder clumps. In typical scenarios, a human operator would respond with additional mixing or a squishing motion to eliminate these clumps. Simulating this behavior in a robotic system is difficult, as it relies on visual detection of clumps, a phenomenon that does not occur consistently. Our research introduces an effective closed-loop control approach to address this challenge, which, to our knowledge, has not yet been explored in existing food industry research.

    \subsection{Foamed Beverages Analysis}

    An important area of focus in food optimization and process automation is the study of powder reconstitution, commonly used in products like coffee, soups, and other beverages~\cite{fang2007measurement}. Specifically, the creation of foam is of high importance to both the drinks industry and consumers~\cite{shweta2020foaming, schuchmann2007product}, with many studies focusing on the optimization and the understanding of the creation and formation of foams. Factors such as water temperature, amount of mixing, pour height, and vessel size all affect the aeration of the beverage and the reconstitution process~\cite{labbe2021impact}, thereby impacting the sensory preference among consumers~\cite{deotale2020foaming}.

    Computer vision has made it possible to automatically assess various quality indicators of foam, including its decay curve~\cite{Cimini2016}, height~\cite{GONZALEZVIEJO2016504}, or the distribution of bubble sizes~\cite{GONZALEZVIEJO2016504, hendriks2020improvement}. Inspired by research in areas such as carbonated beverages~\cite{Viejo2019, BARKER2002565} and flotation froth analysis~\cite{ALDRICH2022107823}, where algorithms like watershed segmentation~\cite{min12091126}, Hough transform~\cite{Romachev2020flotation}, and valley-edge detection~\cite{ALDRICH2022107823} are used for bubble measurement, we developed a specialized computer vision pipeline. This pipeline incorporates three preprocessing approaches specifically designed to analyze bubbles in cappuccino foam. Additionally, we created a foam height measurement algorithm to investigate the relationship between foam height and bubble characteristics.

    \subsection{Experimental Optimization of Food Properties}
    Robotics-driven food preparation optimization remains an under-explored area of research. The variations of Bayesian Optimization (BO) have found applications in food processing optimization~\cite{junge_improving_2020, banga2003improving}. Its ability to efficiently explore complex parameter spaces is particularly well-suited for optimizing food processing tasks that involve multiple interacting variables. Tree-Structured Parzen Estimator (TPE)~\cite{watanabe2023treestructuredparzenestimatorunderstanding} is an alternative black-box method widely used in optimization tasks. However, due to its poor convergence in early experiments, BO was ultimately applied in the final optimization stage, yielding successful results as presented in this study. Specifically, we explore how parameters such as the height of water pouring and the stirring dynamics affect the foam creation and the reconstitution process of powdered beverages. We optimize for the microfoam -- a foam characterized by numerous small bubbles, which improves both the visual appeal and the mouthfeel of the beverage~\cite{huppertz2010foaming}. 

\section{METHODS}
    In this section, we detail the computer vision pipelines, optimization methods, and robotic setup employed to create beverages with varying parameters, evaluate foam quality, and detect undissolved powder.

    \subsection{Coffee Preparation Setup}

\begin{figure}[h]
        \centering
        \includegraphics[width=\columnwidth]{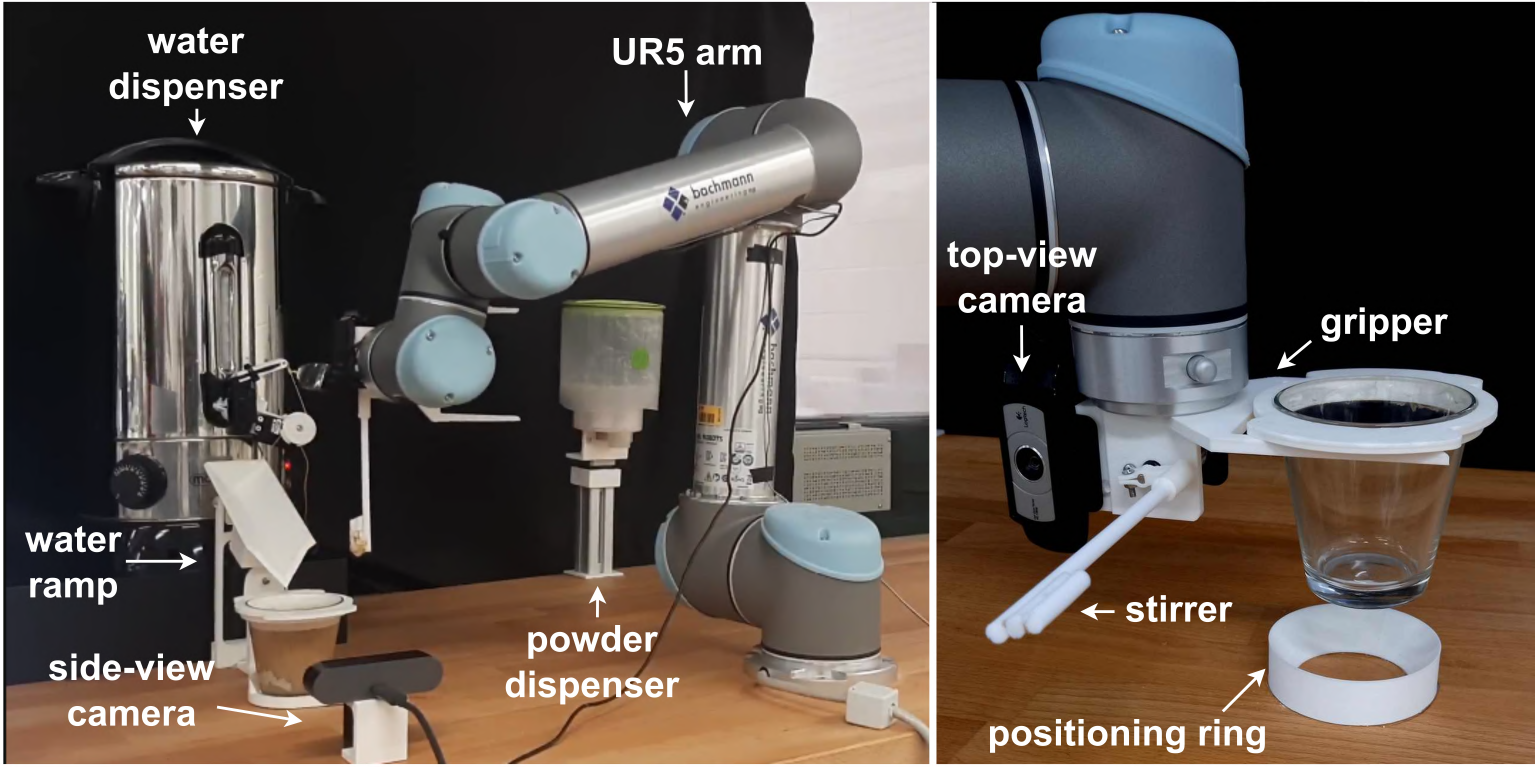}
        \caption{\textbf{Experimental setup}. The powder dispenser, water dispenser, water ramp and robot's end effector were custom-designed and fabricated with the 3D printing technology.}
        \label{fig:setup}
    \end{figure}

    Experimental analysis revealed that the parameters significantly influencing foam formation, while also being easily adjustable, are the height of water pouring ($h$), mixing speed ($s$), and mixing time ($t$). The experimental setup that offers the variability of these parameters is shown in \cref{fig:setup}. A 6 degree-of-freedom UR5 robot is equipped with a custom end effector which allows cups to be moved around. The end effector also features a DC-motor-controlled stirrer and a camera. Transparent cups make the drink easily visible, and a 3D printed rim has been added to the cups for easy and reliable movement. Furthermore, self-aligning cup holders have been designed to ensure cups are placed in a known location.

    To dispense the powder, the cup is moved below a dispensing unit controlled by a stepper motor. The dispenser allows for a fixed quantity of the powder to be poured into the cup. The cup is then moved to the hot water dispenser, whose tap is operated by a servo controlling the duration of the open position period, thereby regulating the volume of added water. A ramp guides hot water from the dispenser to the cup, which allows for the pouring height to be varied with the use of a servo-driven cam mechanism. Water is added while the end effector’s stirrer mixes the content.
    
\begin{figure}[H]
        \centering
        \includegraphics[width=\columnwidth]{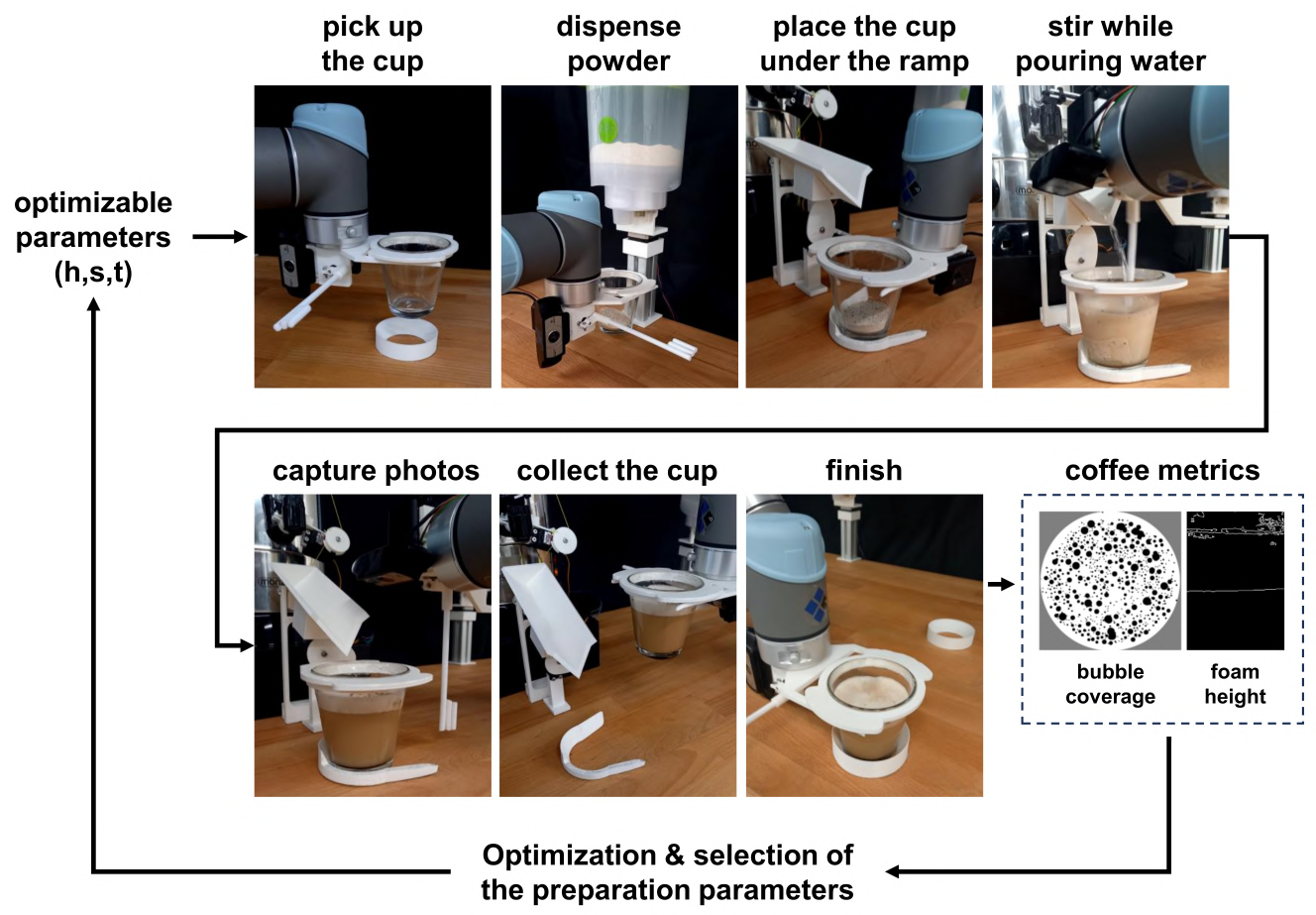}
        \caption{\textbf{Open-loop coffee preparation steps.} This procedure is executed in the optimal parameters search.}
        \label{fig:flowchart}
    \end{figure}

    After mixing is finished, two cameras capture the state of the coffee: one overhead camera mounted on the end effector of the UR5, and the other one fixed on the table to capture the side view of the coffee. An anti-fog coating was applied to the overhead camera to prevent the coffee steam from affecting the image. Once the images are captured, they are analyzed with a computer vision pipeline, and the main controller makes a decision on to how to proceed with each experiment before the coffee is returned to its final location. The flowchart in \cref{fig:flowchart} summarizes the processes and the order of events that take place to make a single coffee.

    \vspace{-0.35cm}    
    The speed of mixing ($s$) corresponds to the speed of the stepper motor, with the range of possible speeds experimentally determined to be between $s=[40\%,100\%]$ of the maximum stepper speed. The time of mixing $t=[0s,60s]$ and the height of the water pouring $h=[10cm,14cm]$ are also adjustable. 

    \subsection{Coffee Analysis}

    To assess the quality of the foam and detect the presence of undesired undissolved clumps of powder a number of computer-vision-based pipelines have been created. It is assumed that both the side view of the transparent cup and the top view of the foam are accessible.

    \subsubsection{Foam Bubbles}
    Bubble assessment is challenging due to the varying size of bubbles, their non-spherical shapes, and the reflective surface of the coffee. To create a robust bubble detection algorithm, three different detection pipelines leveraging blob detection are applied to the same image, and the results are then combined. Ideally, in the case of microfoam, no bubbles would be visible to this computer vision system.

\begin{figure}[h]
        \centering
        \includegraphics[width=\columnwidth]{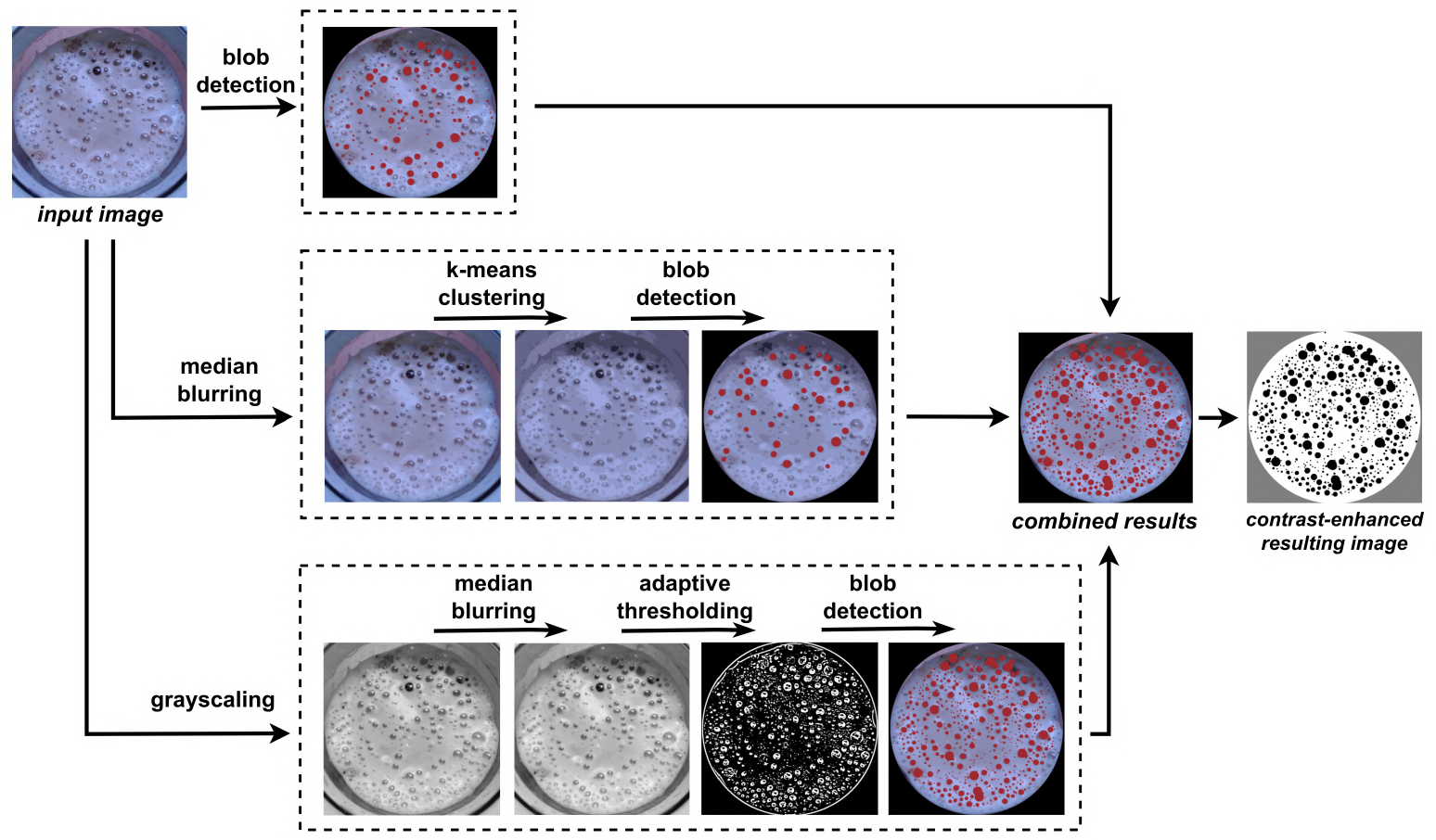}
        \caption{\textbf{Bubble coverage determination pipeline.} The results of three simultaneous processes are combined to identify the bubble coverage.}
        \label{fig:bubbles}
    \end{figure}

    The first pipeline directly identifies small bubbles on the input image. The second one applies preprocessing with median blurring and K-means clustering to identify larger and non-spherical bubbles. The third detector uses grayscale conversion, median blurring, and adaptive thresholding, which is particularly effective for larger bubbles or those with reflections. The blobs detected with each of the three pipelines are then combined into an single black-and-white image, where the percentage area of blobs is determined by totaling the area of black pixels. This approach is summarized in \cref{fig:bubbles}.

\begin{figure}[h]
        \centering
        \includegraphics[width=\columnwidth]{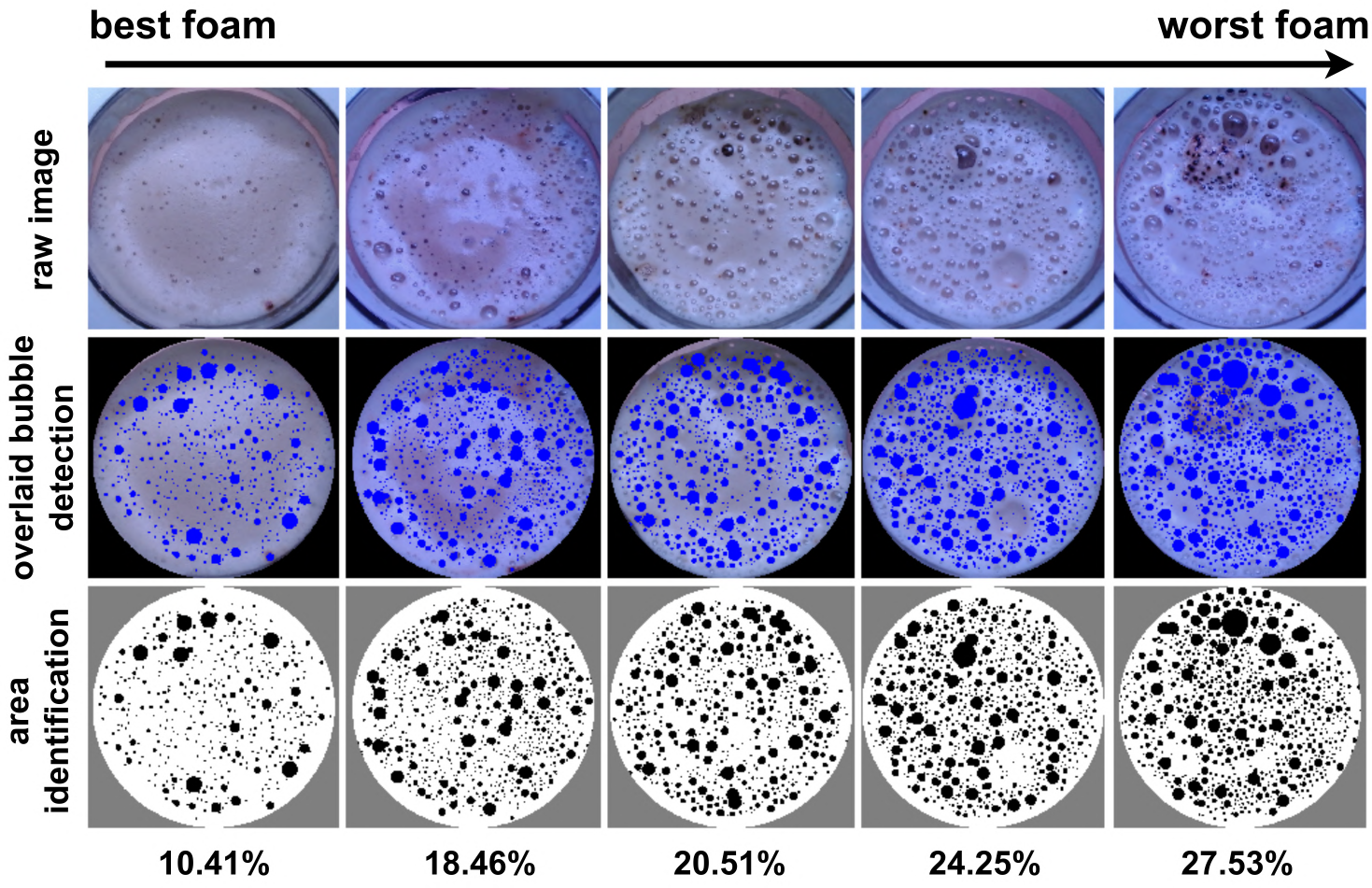}
        \caption{\textbf{Bubble detection for foams of variable quality}. This overview demonstrates that bubble coverage is an effective and reliable metric of the foam quality.}
        \label{fig:foam_quality}
    \end{figure}

    To demonstrate how this proposed method provides a representative metric for foam quality, coffees with foams of varying quality were prepared. As shown in \cref{fig:foam_quality}, the best foam has a very low area of bubbles ($10.41\%$), whereas the worst foam has an area of 27.53\%. Within this range, the area metric increases monotonically with the decrease of the quality of the foam. This indicates that the metric corresponds to the visual quality of the foam and provides significant differentiation to capture the varying quality of foams.

    \subsubsection{Foam Height}
\begin{figure}[h]
        \centering
        \includegraphics[width=\columnwidth]{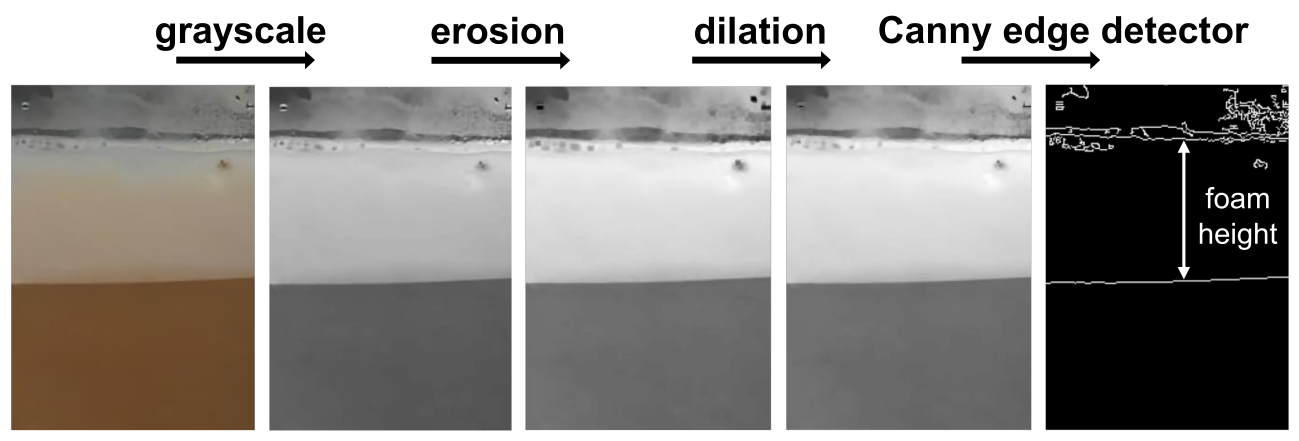}
        \caption{\textbf{Foam height determination pipeline}. The analysis is performed using a side-view image of the cup.}
        \label{fig:foam_height}
    \end{figure}

    Foam height is a second metric used to define foam quality. Accurately assessing the foam height is challenging because, in the side view, the top of the foam can be difficult to see due to condensation on the glass and the presence of bubbles on the surface. To measure the foam height, the image is first converted to greyscale with an erosion and dilation applied, after which a Canny edge detector is used to identify the edges corresponding to the bottom and the top of the foam. The mean difference between these edges provides an estimation of the foam height. The approach is summarized in \cref{fig:foam_height}.

    \subsubsection{Clump Detection}
\begin{figure}[H]
        \centering
        \includegraphics[width=\columnwidth]{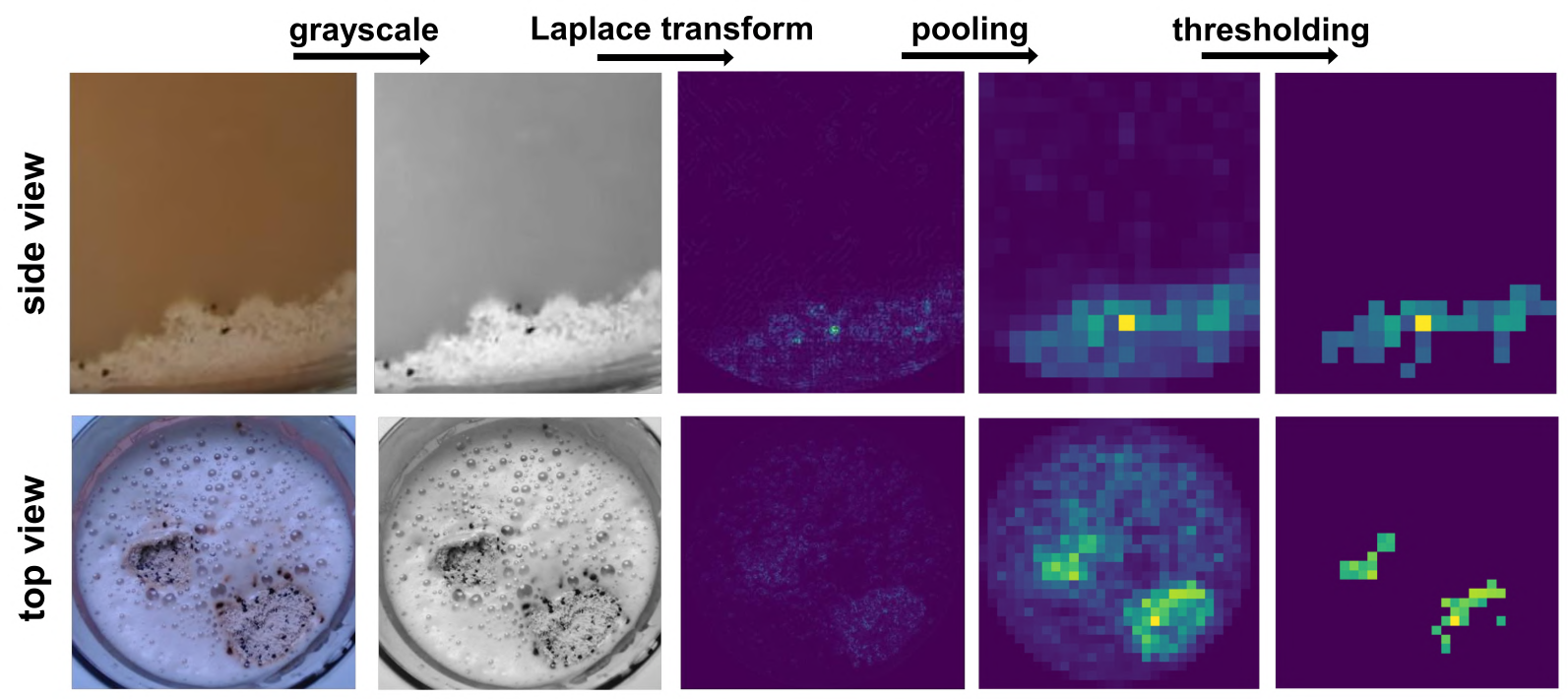}
        \caption{\textbf{Clump detection pipeline}. Undissolved powder clumps may be present both within the foam and at the bottom of the cup.}
        \label{fig:clumps}
    \end{figure}

    This detection focuses on the presence of undissolved powder clumps in both the foam and the bottom of the cup. A robust approach applicable for both of these cases has been developed. The images are first converted to greyscale, followed by an application of a Laplace transform. Then, a customized pooling returns a matrix filled with sums of absolute values of framed pixels. By thresholding pixel values, the presence and approximate area of the clumps can be determined.  A demonstration of this approach is shown in ~\cref{fig:clumps}.

    \vspace{-0.5cm}

    \subsection{Closed-loop Clump Removal}

    To simulate consumer behavior in clump removal, the size of the clump (in pixels) is first detected ($c$). Based on it, a proportional controller in the form of $t_m = \alpha c$ determines the additional mixing time $t_m$ required to remove the clump with the stirrer, either in the foam or at the bottom of the cup. Through heuristic testing, the optimal value $\alpha$ of the controller was determined to be 0.2, representing the best result found during experimentation. It typically results in mixing times ranging from 5 to 20 seconds, depending on the size of the clump.

    \subsection{Coffee Optimization}

    Let us consider the objective function $f:\mathcal{X} \rightarrow \mathbb{R}$, which measures the quality of the coffee as the percentage of area without bubbles visible in the coffee foam. $\mathcal{X}$ denotes a bounded domain $\mathcal{X} = [0,60] \times [40,100] \times [10,14]$. A point $\textbf{x} \in \mathcal{X}$ is expressed as $\textbf{x} = (t,s,h)$, where $t$ [s] is mixing time, $s$ [\%] is mixing speed, $h$ [cm] is water pouring height. The goal is to maximize the function $f$ over the bounded domain, i.e. we want to find $\arg \max_{\textbf{x} \in \mathcal{X}} f(\textbf{x})$ to effectively optimize for microfoam. As $f(\textbf{x})$ is unknown, Bayesian Optimization (BO), recognized as one of the most efficient sampling algorithms for black-box functions~(\cite{nguyen2019bayesian}), was selected as the optimization method. It is particularly effective when only a few parameters need to be optimized~(\cite{saar2018modelfree}). With BO, a Gaussian Process  $\mathcal{GP}$ prior is placed on $f(\textbf{x})$:
    $$
    f(\textbf{x}) \sim \mathcal{GP}(m(\textbf{x}), k(\textbf{x}, \textbf{x}')),
    $$
    where $m(\textbf{x})$ is the mean function, in this case set to zero, and $k(\textbf{x}, \textbf{x}')$ is a covariance kernel being a Matern kernel with $\nu=2.5$.

    At iteration $n$, $n$ distinct points $\{x^{(i)}\}^n_{i=1} \subset \mathcal{X}$ have been observed, with corresponding values   $\{y^{(i)}\}^n_{i=1}$, where $y^{(i)}=f(\textbf{x}^{(i)})+\epsilon^{(i)}$, with $\epsilon^{(i)}$ accounting for observational noise. Given these data, the posterior predictive distribution for any new point is a normal distribution:
    $$
    f(\textbf{x}) | \{x^{(i)},y^{(i)}\}^n_{i=1} \sim \mathcal{N}(\mu_n(\textbf{x}), \sigma^2_n(\textbf{x})),
    $$
    where $\mu$ and $\sigma$ respectively correspond to mean and variance, whose parameters are fitted to the data by maximizing the Gaussian Process' log marginal likelihood after each observation.

    To decide the next experimental point $\textbf{x}_{n+1}$, an Upper Confidence Bound acquisition function was selected:
    $$
    \alpha_{n}(\textbf{x}) = \mu_n(\textbf{x}) + \kappa \sigma_n(\textbf{x}),
    $$
    where $\kappa$ is a parameter for controlling the exploration-exploitation tradeoff. Hence, at step $n$, the next point $\textbf{x}_{n+1}$ is chosen as:
    $$
    \textbf{x}_{n+1} = \arg \max_{\textbf{x} \in \mathcal{X}} \alpha_n(\textbf{x})= \arg \max_{\textbf{x} \in \mathcal{X}} [\mu_n(\textbf{x}) + \kappa \sigma_n(\textbf{x})].
    $$
    The value of $\kappa=8$ was chosen, reflecting the need to explore the design space before exploiting and finding the optimal solution.
    
    The more observations provided to the optimizer, the more confident the algorithm becomes regarding its prediction of optimal parameters. Despite the efficiency of this approach, searching a design space that has three parameters that exhibit variance and stochastic nature of the results requires tens or hundreds of trials to form an accurate model that can also provide a solution with a good performance. 

\section{RESULTS}

\subsection{Coffee Preparation \& Optimization}
    To demonstrate the coffee making process and evaluate its repeatability, four cappuccinos were automatically prepared with the same input conditions. The results indicate a variability of 0.8\%. While the localization of the bubbles on the foam surface varies, there are clear similarities in the density and size of the bubbles present. This highlights the need for larger-scale physical experiments and for the use of BO in the optimization processes, as it can handle this variability.

\begin{figure}[h]
        \centering
        \includegraphics[width=\columnwidth]{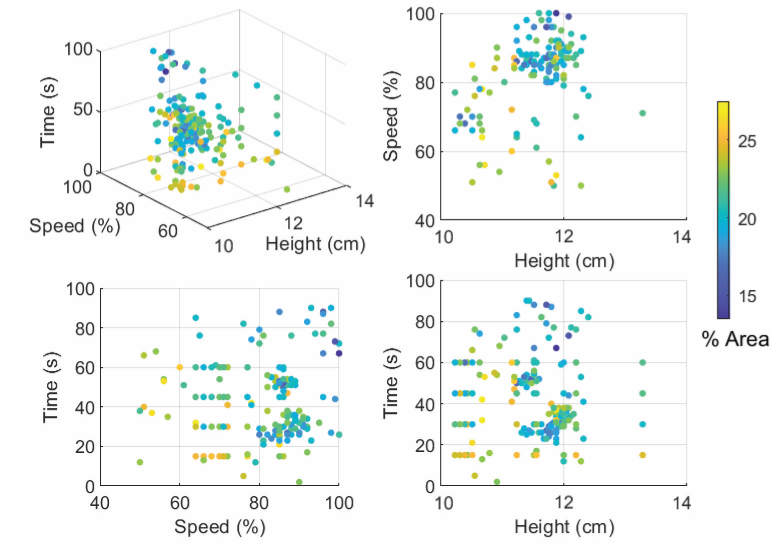}
        \caption{\textbf{Bubble coverage results.} The plots display data from over 100 coffees, prepared with varying stirring speed, stirring time and water pouring height.}
        \label{fig:coffee_experiments}
    \end{figure}

    Before performing BO, approximately 100 coffees were made using the experimental setup. This included a mix of grid-based and random exploration to investigate the design space and the observed variability. The bubble area of these coffees as a function of the mixing speed, time and pour height is shown in \cref{fig:coffee_experiments}. The results demonstrate the complexity of the interactions. Low pouring height and quick mixing at high speeds result in some instances of low bubble coverage. However, other parameter sets also produce high quality coffees. This further supports the use of BO, as there is no single local minima that can be found with simple gradient descent-based methods.

    Fifty coffees were prepared with the objective function of BO set to minimize the bubble coverage. The results in~\cref{fig:bo} show that although there are fluctuations in the optimization’s exploration, the process converges to a minimum value over time, with a minimal bubble area of around 11\% found. This corresponds to a low water height (11cm), low mixing speed (65\%) and a high mixing time (50 s). From around 45 iterations onwards, these values remain approximately constant with limited further exploration, particularly in the case of the mixing speed.

    \begin{figure*}[h]
        \centering
        \begin{subfigure}{0.35\textwidth}
            \includegraphics[width=\textwidth]{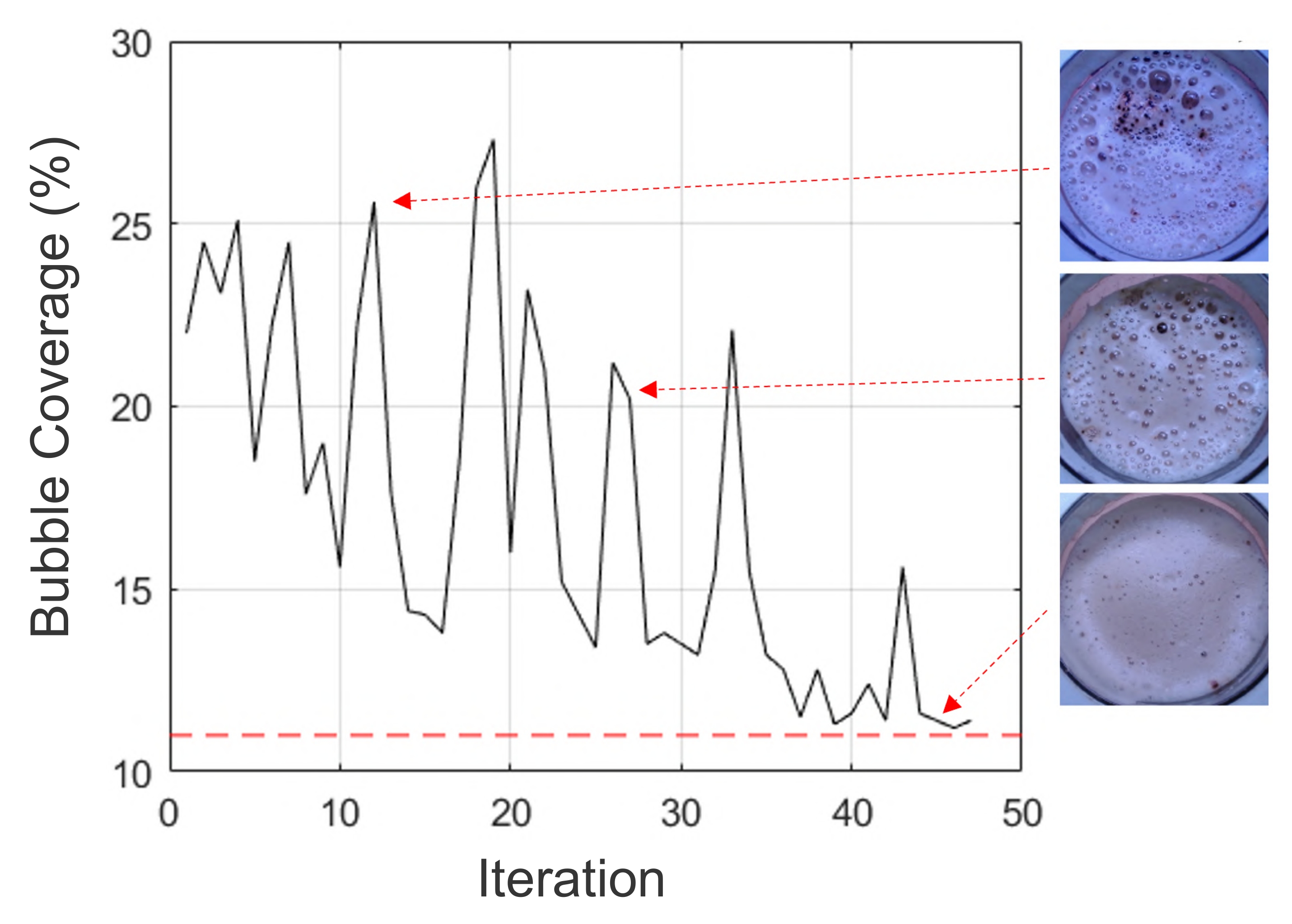}
            \caption{Bubble coverage optimization}
            \label{fig:bo}
        \end{subfigure}
        ~
        \begin{subfigure}{0.29\textwidth}
            \includegraphics[width=\textwidth]{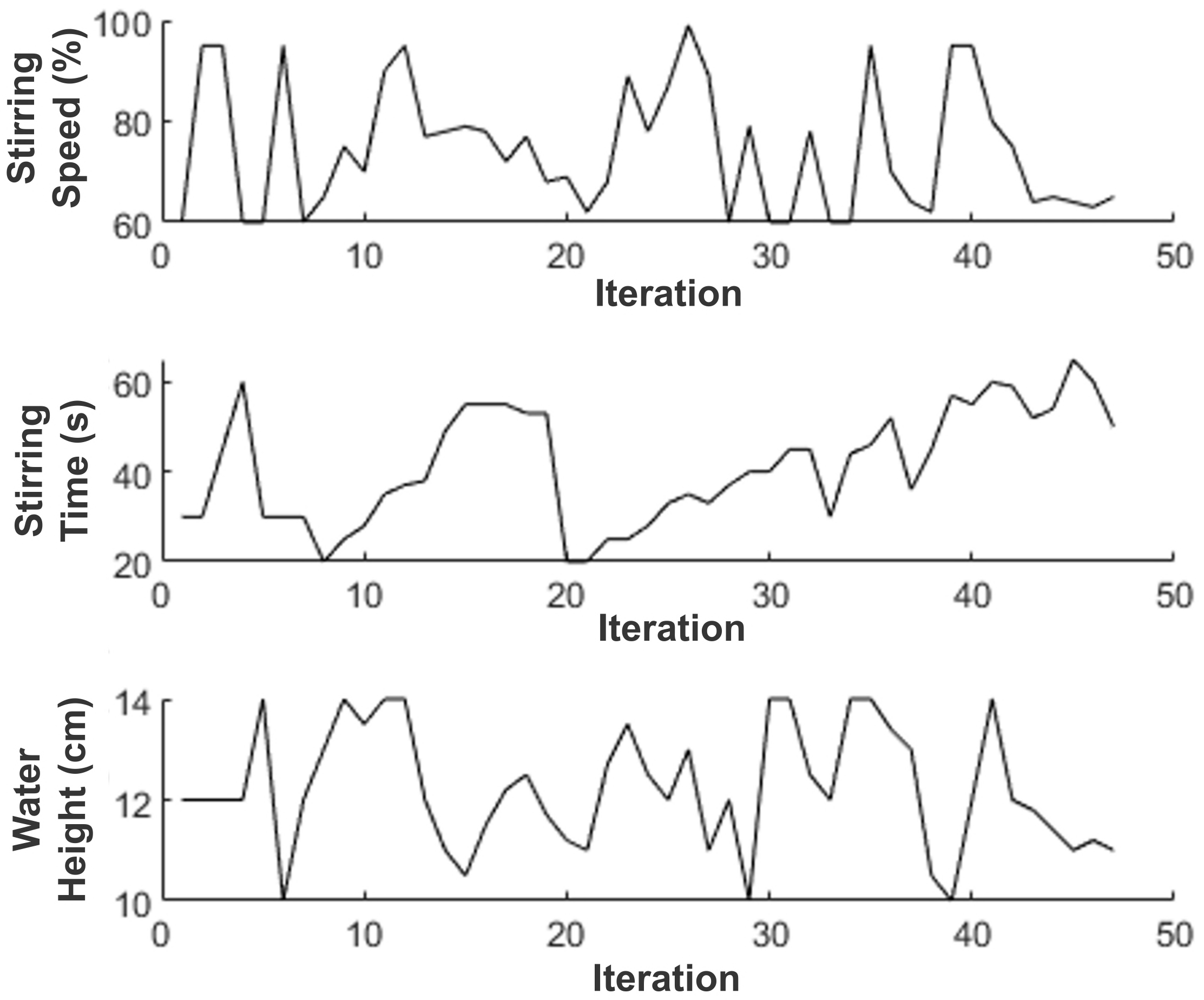}
            \caption{Parameter exploration}
            \label{fig:bo_other}
        \end{subfigure}
        ~
        \begin{subfigure}{0.3\textwidth}
            \includegraphics[width=\textwidth]{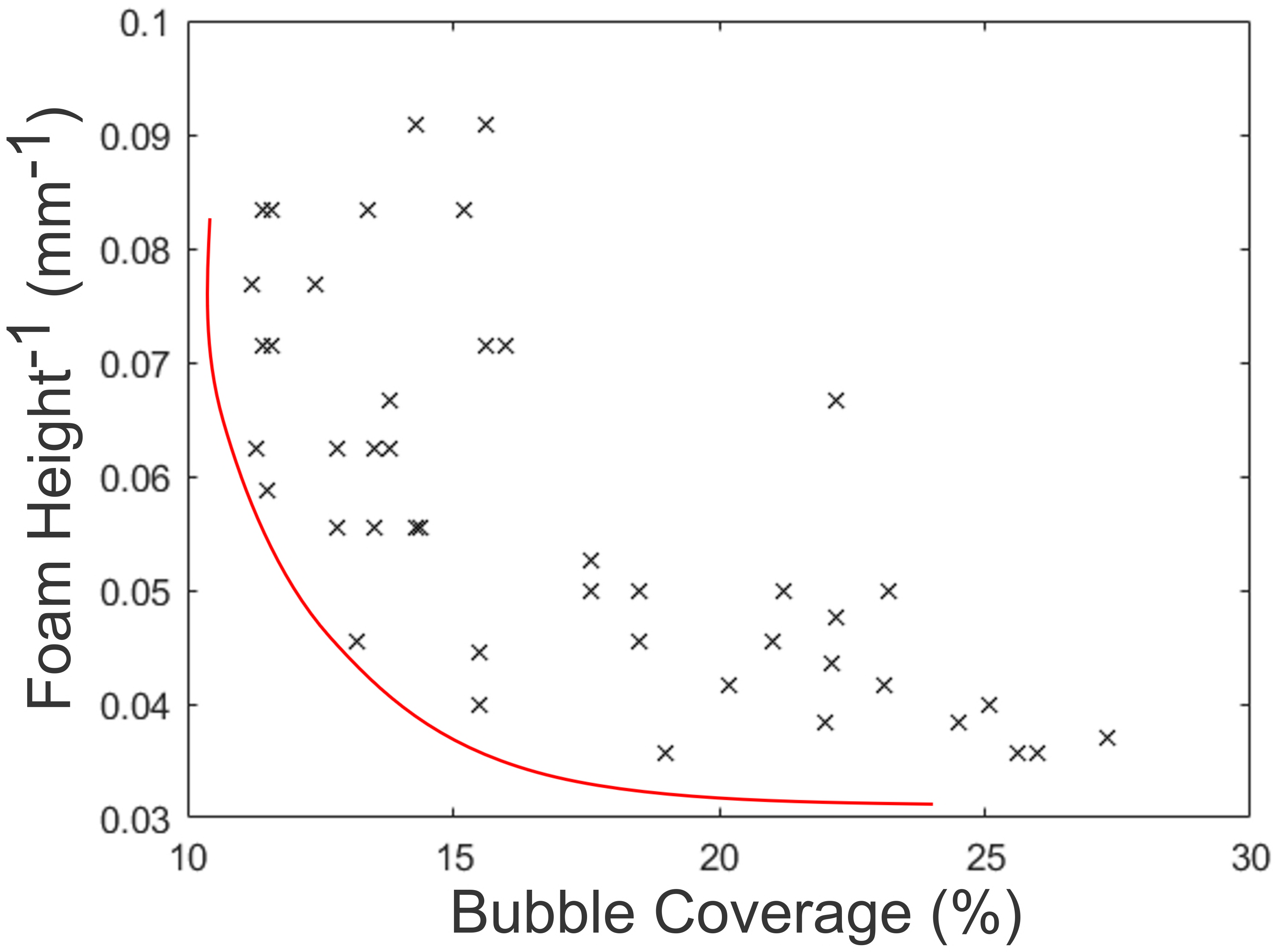}
            \caption{Output metrics relationship}
            \label{fig:pareto}
        \end{subfigure}
        \caption{{\bf Bayesian Optimization results.} Following approximately 50 iterations of the optimization process, the optimal parameters were successfully identified.}
        \label{fig:bo}
    \end{figure*}

\begin{figure}[h]
        \centering
        \includegraphics[width=0.75\columnwidth]{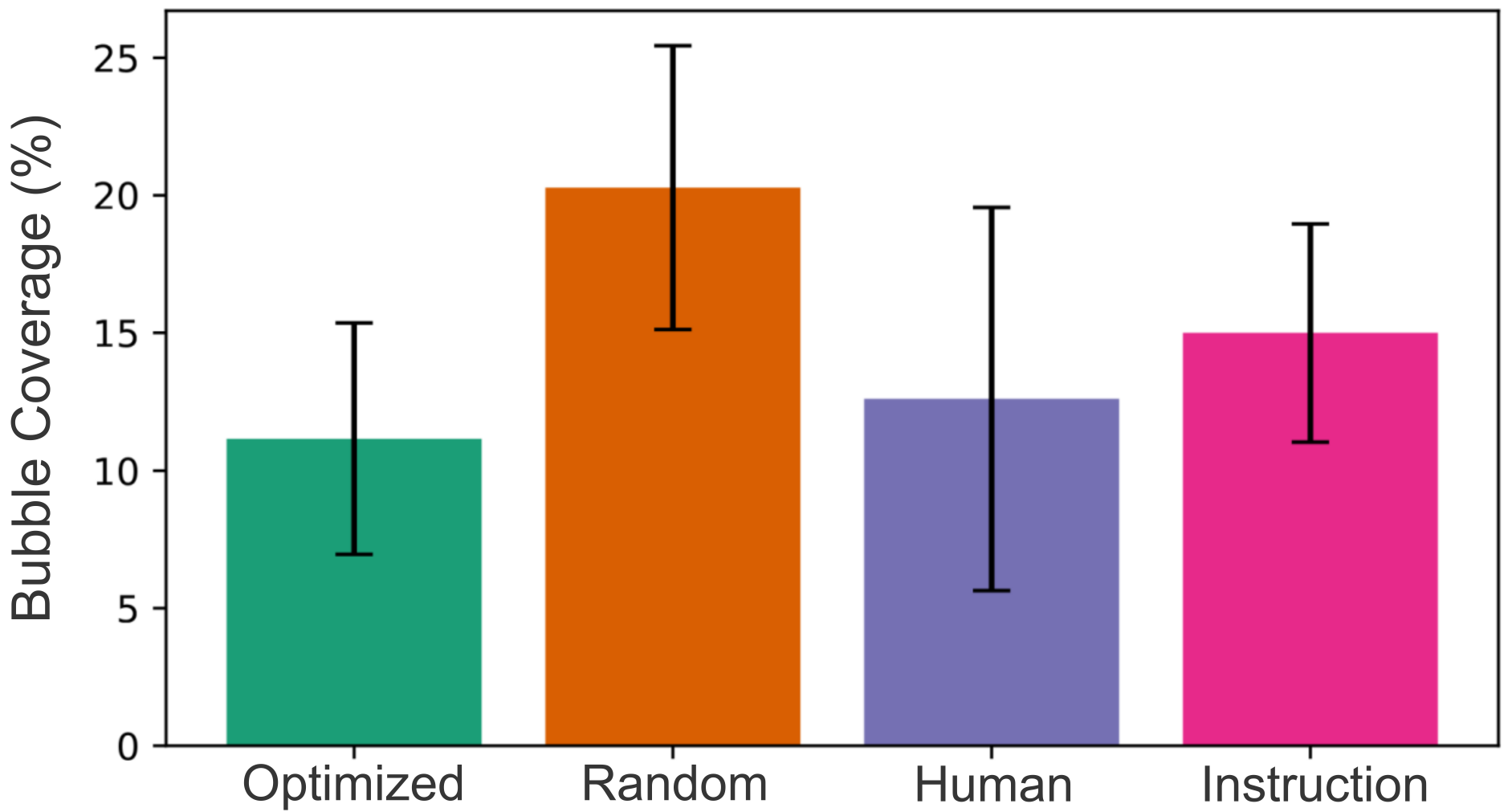}
        \caption{\textbf{Comparison of the best results achieved by different methods}. The plot presents the resulting average bubble coverage along with the standard deviation.} %
        \label{fig:comparison}
    \end{figure}

    As BO provides no guarantee of optimality, the optimal processing conditions found were compared to others: 1) random selection of mixing conditions, 2) coffee preparation by a human, 3) robot following the provided producer’s instructions. For each of these cases, four coffees were made with the results presented in~\cref{fig:comparison}, proving that BO-defined conditions outperformed the others. Although the human-made coffees had on average only 1\% more bubble coverage compared to BO, they showed higher variance. Given that the human can use continuous visual feedback to adaptively mix the coffee, this highlights the quality of the optimal performance found using BO. The random preparation parameters presented the worst performance, and the instruction-based performance was inferior to the one of a human. 
    
    We additionally investigated the relationship between bubble coverage and foam height. For each of the 50 coffees prepared in the BO experiment, the foam height against the bubble coverage is plotted~\cref{fig:pareto}. Interestingly, this showed that an increase in the bubble area is followed by an increase in the height of the foam. Therefore, there is a potential trade-off between foam height and bubble area, showing a Pareto-optimality problem. This requires further exploration in the future research.

    \subsection{Closed-loop Clump Removal}
\begin{figure}[h]
        \centering
        \includegraphics[width=0.95\columnwidth]{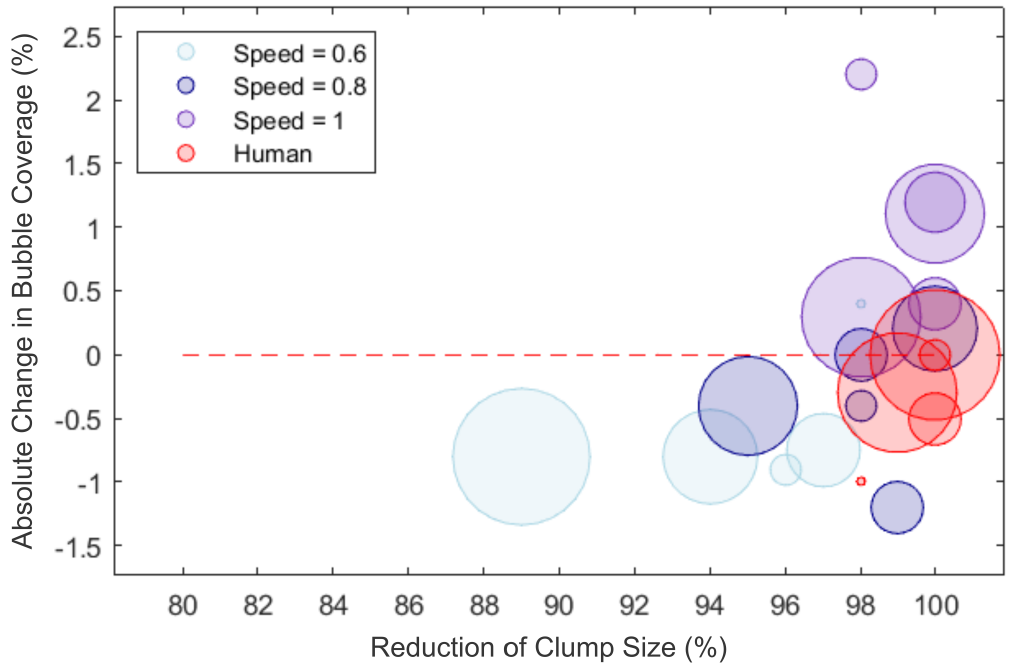}
        \caption{\textbf{Results of the closed-loop clump reduction with the proportional controller}. The plot illustrates the reduction in clump size and the absolute change in bubble coverage, with bubble radius in the plot representing clump size. The dotted line indicates zero change in bubble area.}
        \label{fig:clump_res}
    \end{figure}

    This experiment examined the ability to detect and remove clumps, how this process affects the foam bubble area, and how similar the robot’s closed-loop removal is to human behavior. The coffees were prepared with the optimal parameters found with BO. The bubble area was recorded, and if there was a presence of clumps, their area was identified. The proportional controller was then used with varying mixing speeds ($60\%$, $80\%$ and $100\%$). The results from the twenty experiments with the clump-containing cappuccinos are presented in~\cref{fig:clump_res}, showing the reduction in the clump size (i.e. the success of clump removal) and the change in bubble coverage area. For comparison, a human was asked to remove the clumps with a spoon, and the same metrics were recorded. As the occurrence of clumps is hard to reproduce on purpose, the clump size varied every time.

    The ideal removal procedure should fully remove the clump and either reduce the bubble area or have minimal effect on it. Although high speed mixing removed the clumps, it had a negative impact on the foam by increasing the bubble coverage. Conversely, a low speed resulted in less success in removing the clumps, but reduced the bubble area. The speed of 80\% of the maximum showed behavior most similar to that of humans, where the clumps were mostly reduced, but the change in bubble area is low. Interestingly, although the human was very good at removing the clumps, the bubble area did not decrease.
    
    The results in this section demonstrate that similar behavior to humans can be achieved with the robotic setup. Additionally, the proportional controller proved to be a suited tool for the effective removal of clumps, even contributing to the improvement of foam quality.

\section{DISCUSSION}

    In this work we presented an automated approach to optimize the foam of reconstituted beverages by leveraging robotics, computer vision, and optimization algorithms. We demonstrated how this can facilitate large-scale experiments under controlled conditions, with an application in optimizing foam quality through 50 iterations. Furthermore, we showed that a computer vision-aided feedback loop can simulate human behavior in clump removal and how this process affects the resulting properties of the foam.

    Although this work focuses on foam optimization, it highlights the role that robotics and computer vision can play in food science, identifying optimal process parameters combinations. This is increasingly important for developing products optimized for nutrition inclusion, cost, and processability. The use of computer vision provides a non-contact means of analyzing food and drinks. It is cost-effective compared to many standard analysis methods and shows versatility; in our case, it can be used to analyze various aspects of the foam.
    
    These results open up many interesting directions for future investigation. Exploring a larger number of parameters (e.g., water temperature or stirrer size) would further exploit the potential of robotics for recording measurements and automatic task execution. Additionally, research on more complex fitness functions, such as minimizing the presence of clumps, could extend the applicability of Bayesian Optimization. Finally, exploring alternative learning-based approaches, particularly suited to image-based analysis, could offer a promising solution for generalizing the optimization approach – for example, by providing a food or drink sample with specific qualities that the robot should seek to achieve. In summary, the use of robotics and computer vision opens up many exciting directions in food science, which should be leveraged from product development to the creation of consumer instructions.

\section*{ACKNOWLEDGMENTS}
We extend our sincere gratitude to Cecile and other Nestle Research colleagues for their invaluable support and contributions to this research.

\newpage{\pagestyle{empty}\cleardoublepage}

\end{document}